\pdfoutput=1

\documentclass[11pt]{article}

\usepackage{coling}

\usepackage{times}
\usepackage{latexsym}

\usepackage[T1]{fontenc}

\usepackage[utf8]{inputenc}

\usepackage{microtype}

\usepackage{inconsolata}

\usepackage{graphicx}

\usepackage{times}
\usepackage{latexsym}
\usepackage{algorithm, algorithmic}
\usepackage{amsmath} 
\usepackage{amssymb}
\usepackage{multirow}
\usepackage{arydshln} 
\usepackage{makecell}
\usepackage{hhline}
\usepackage{xspace}
\usepackage{booktabs}
\usepackage{float}
\usepackage{graphicx}

\usepackage{amssymb}
\usepackage{pifont}
\usepackage{bm}
\usepackage{enumitem}
\usepackage{diagbox}
\setlist[itemize]{leftmargin=*}
\setitemize[1]{itemsep=0pt,partopsep=0pt,parsep=\parskip,topsep=0pt}
\usepackage{microtype}
\title{LLMCL-GEC: Advancing Grammatical Error Correction with LLM-Driven Curriculum Learning}

\author{
        Tao Fang$^1$~~~~
        Derek F. Wong$^1$\thanks{~~Corresponding Author}~~~~ 
        Lusheng Zhang$^2$~~~~
        Keyan Jin$^3$~~~~
        Qiang Zhang$^{2 4}$~~~~\\
        \textbf{Tianjiao Li}$^2$~~~~
        \textbf{Jinlong Hou}$^2$~~~~
        \textbf{Lidia S. Chao$^1$}~~~~
         \\
    $^1$NLP$^2$CT Lab, Department of Computer and Information Science, University of Macau \\
      \texttt{nlp2ct.taofang@gmail.com~~\{derekfw,lidiasc\}@um.edu.mo} \\
    $^2$Bilibili Inc., LLM Team~~~~$^3$Macao Polytechnic University~~~~$^4$Xi'an Jiaotong University \\
      \texttt\{Zhanglusheng01, litianjiao, kinghou\}@bilibili.com~~ \\
    \texttt{p2317001@mpu.edu.mo}~~ \\
    \texttt{zhangqiang@stu.xjtu.edu.cn}
    }

\begin{document}
\maketitle
\begin{abstract}
While large-scale language models (LLMs) have demonstrated remarkable capabilities in specific natural language processing (NLP) tasks, they may still lack proficiency compared to specialized models in certain domains, such as grammatical error correction (GEC). Drawing inspiration from the concept of curriculum learning, we have delved into refining LLMs into proficient GEC experts by devising effective curriculum learning (CL) strategies.  In this paper, we introduce a novel approach, termed LLM-based curriculum learning, which capitalizes on the robust semantic comprehension and discriminative prowess inherent in LLMs to gauge the complexity of GEC training data. Unlike traditional curriculum learning techniques, our method closely mirrors human expert-designed curriculums. Leveraging the proposed LLM-based CL method, we sequentially select varying levels of curriculums ranging from easy to hard, and iteratively train and refine using the pretrianed T5 and LLaMA series models. Through rigorous testing and analysis across diverse benchmark assessments in English GEC, including the CoNLL14 test, BEA19 test, and BEA19 development sets, our approach showcases a significant performance boost over baseline models and conventional curriculum learning methodologies.
\end{abstract}

\begin{figure}[!t]
\centering
\includegraphics[width=0.45\textwidth, trim=0 0 0 0]{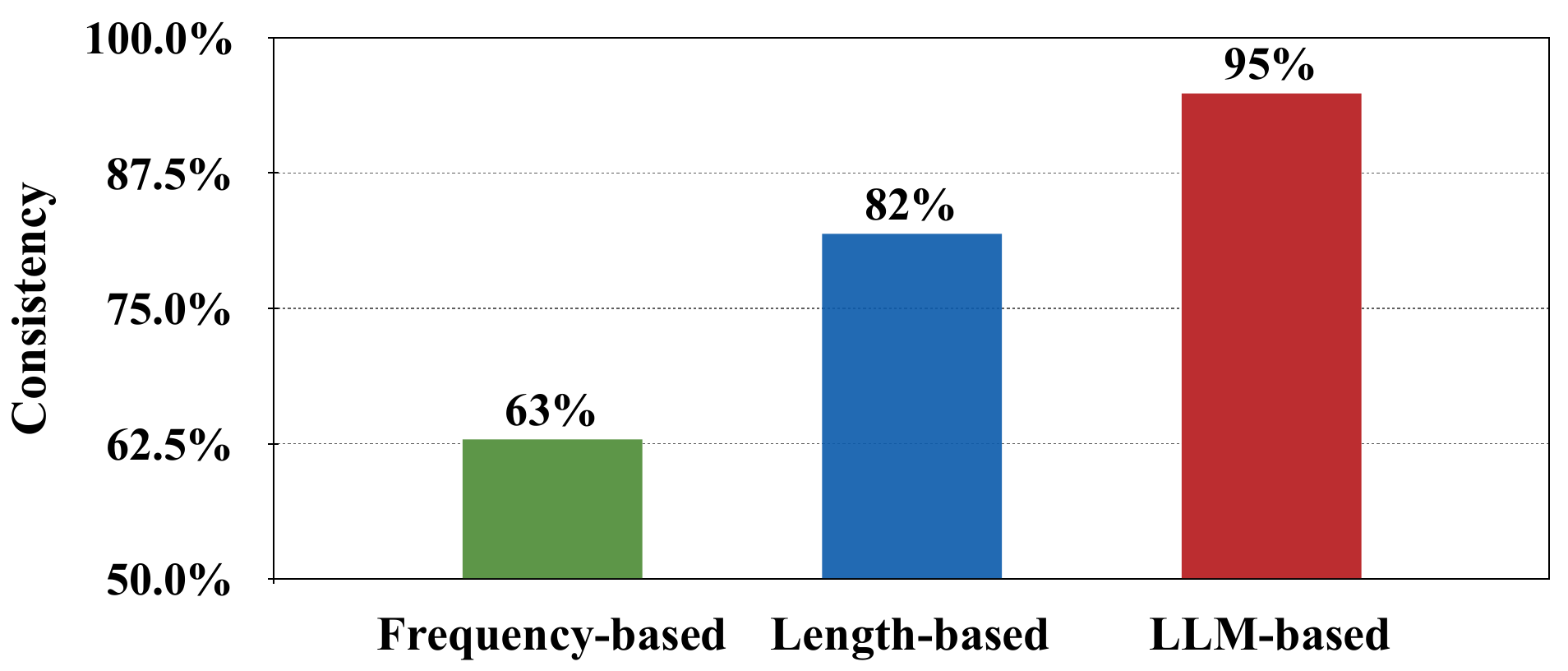}
\caption{Different curriculum learning methods exhibit varying degrees of consistency with the standards set by human experts in terms of establishing the difficulty levels of data. Our proposed \textbf{LLM-based CL} approach demonstrates a high level of alignment with these expert-defined criteria.}
\label{fig:freq_len_llm}
\end{figure}

\section{Introduction}
Large Language Models (LLMs) have garnered considerable attention due to their exceptional performance across a wide array of downstream Natural Language Processing (NLP) tasks \cite{NEURIPS2020_1457c0d6,touvron2023llama,touvron2023llama2,openai2024gpt4}. Notably, the deployment of LLMs has been particularly influential in the fields of text continuation, general dialogue, and an assortment of other text generation tasks \cite{jiao2023chatgpt,hendy2023good,pan2023preliminary,fang2023chatgpt,loem-etal-2023-exploring}.
However, 
the evaluation of LLMs in downstream NLP tasks presents a complex scenario. Although LLMs demonstrate certain benefits, they also manifest a noticeable performance deficit in comparison to specialized models, particularly in tasks such as grammatical error correction (GEC) \cite{fang2023chatgpt,loem-etal-2023-exploring}. 
The proposed solutions, including prompt adjustments and supervised fine-tuning on fixed datasets \cite{zhang2023multitask}, tend to resolve immediate performance issues without substantially enhancing the model's intrinsic understanding of the task. 
This predicament can be likened to a highly learned professor excelling in their field of expertise but having a superficial understanding of other disciplines. 
To acquire proficiency in new areas, such a professor would employ a structured learning approach, incrementally developing their expertise. 
Similarly, for LLMs to master downstream tasks, they necessitate a structured learning curriculum that sequentially progresses from easy to hard. 

This curriculum learning (CL) approach \citep{10.1145/1553374.1553380,ren2019learning,data-parameters} not only lays a solid knowledge foundation but also equips the model to adapt and refine its understanding with increasing task complexity, ultimately enhancing performance and applicability across NLP tasks. 
Researchers often establish different curriculum guidelines for various NLP tasks. For example, in machine translation (MT) tasks, considerations typically include sentence length and the frequency of rare words to determine the curriculum's difficulty level \cite{kocmi-bojar-2017-curriculum,platanios-etal-2019-competence,liu-etal-2020-norm}. However, to date, we have not seen researchers employ CL methods for GEC tasks, particularly within LLMs. Considering the similarities between GEC and MT tasks \cite{yuan-briscoe-2016-grammatical,zhou-etal-2020-improving-grammatical}, we believe that establishing a curriculum based on sentence length for GEC tasks can be widely accepted. However, there is a practical issue: not all short sentences are easy samples, and not all long sentences are hard samples. For instance, translating short sentences with nuanced meanings is not straightforward, while a completely correct long sentence may require minimal effort for the model. Based on this, we believe that effective curriculum learning should be guided by human experts to set the CL difficulty level. However, this approach incurs a high cost in terms of annotating tens of thousands of training samples. Fortunately, numerous studies have demonstrated that LLMs can attain human-level performance in comprehension and discrimination \cite{touvron2023llama2,openai2024gpt4,minaee2024large}. Moreover, extensive efforts have been dedicated to employing LLMs for data selection and generation \cite{albalak2024survey,xia2024less}. Motivated by these findings, we opt to utilize LLMs to devise a curriculum for the GEC task. Figure~\ref{fig:freq_len_llm} illustrates the consistency of different difficulty levels of courses set by LLMs and two other common curriculum learning methods with those set by human experts (200 randomly selected sentences from the cLang8 training data \cite{rothe-etal-2021-simple}). Notably, the curriculum devised by LLMs exhibits a strong consistency with that established by human experts.

In this paper, we introduce a novel curriculum learning method for the GEC task, termed the LLM-based curriculum learning method. This approach leverages a large LLM (LLaMA2-70b) \cite{touvron2023llama2} as an expert to assess the correction difficulty of English GEC source-side training data and then devises various courses (easy->medium->hard) based on the scores provided by the LLM expert. We conduct iterative training from easy to hard courses using the curriculum designed by the LLM expert for the English GEC task on different model types. The experimental results and analysis on the CoNLL14 test \citep{ng-etal-2014-conll}, as well as the BEA19 test and development sets  \citep{bryant-etal-2019-bea}, demonstrate significant improvements over baseline models and other curriculum learning methods across both the T5 series and LLaMA series models. In addition, we explore the sequencing of training in curriculum learning forms. Experimental results indicate that models derive benefits only from a learning process that progresses from easy to hard; conversely, learning in reverse is nearly ineffective, thus reaffirming the efficacy of curriculum learning in GEC. Compared to models trained all at once with the entire training data, we update the baseline by summing up the data iteratively trained in curriculum learning and use it to train models all at once for comparison. We find that only by adhering to the principle of curriculum learning and progressively incorporating difficult samples could significantly enhance the GEC model performance.

\begin{figure*}[!t]
\centering
\includegraphics[width=0.88\textwidth, trim=0 0 0 0]{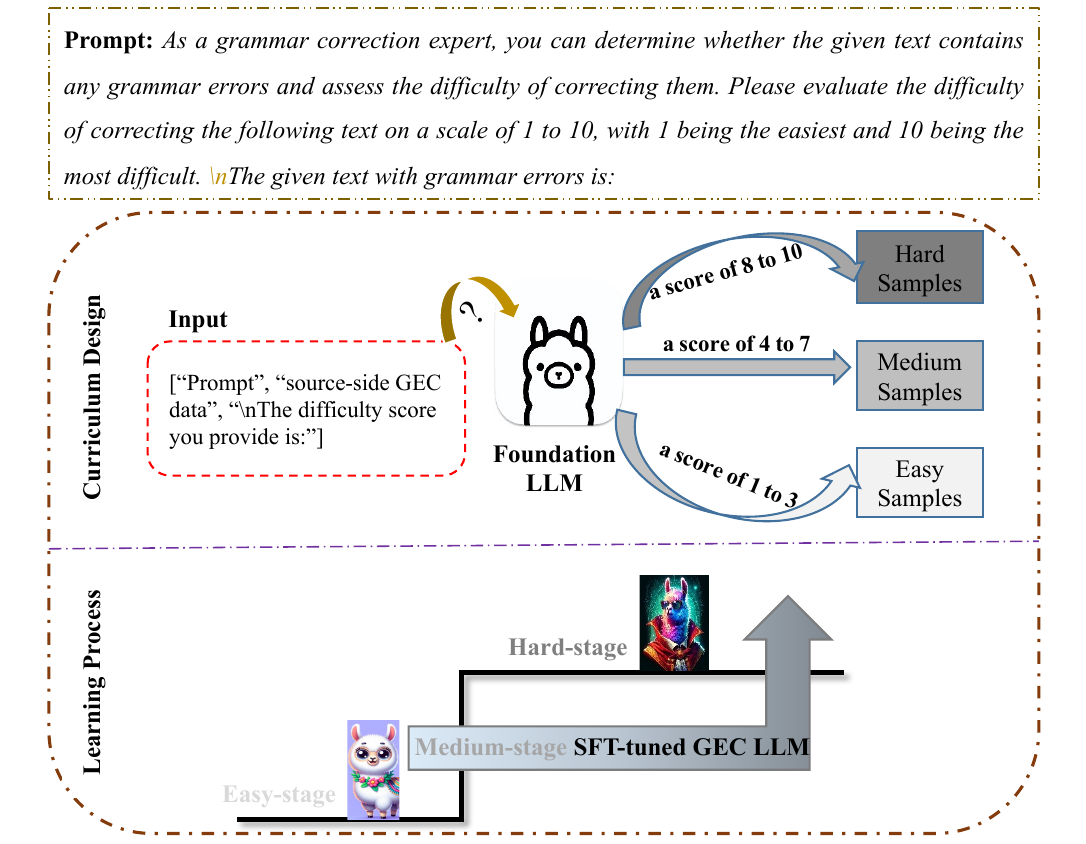}
\caption{The overall framework of our proposed LLM-based CL method consists of two parts: one involves using a foundation LLM (LLaMA2-70b) to establish courses of different levels, while the other involves following the principles of curriculum learning to perform SFT on the GEC model using the established curriculums.}
\label{fig:cl_method}
\end{figure*}

\section{Method}
Our primary focus revolves around constructing an adaptable curriculum learning framework and leveraging the curated curriculum data to train LMs into adept specialists for downstream tasks. Our methodology comprises two key phases: Firstly, employing an LLM to assess the difficulty level of the training data, followed by categorizing the scored data into three tiers: easy, medium, and hard courses. Secondly, sequentially exposing the LM to these curriculum tiers for training-commencing with easier samples, progressing to medium difficulty, and culminating with challenging material ultimately achieving proficiency across the entire learning process. Figure~\ref{fig:cl_method} illustrates the procedural framework of our proposed approach.
\subsection{Curriculum Design}
\paragraph{Large-scale Language Model} The pre-training of large-scale language models relies on a vast amount of unsupervised data \cite{touvron2023llama,touvron2023llama2,openai2024gpt4}. Assuming a sentence $X$ consists of $N$ tokens $x_{1},...,x_{n}$, and a context window of size $k$, the training objective of the LLM is:

\begin{equation}
 \underset{\theta}\arg \max p(X;{\theta})  = \sum \log p_{\theta}(x_{i} | x_{i-k:i-1}),
\end{equation}
where $\theta$ denotes the trainable parameter.

\paragraph{Curriculum Construction} LLMs that have completed training typically possess strong semantic comprehension and emergent capabilities. They can be guided to complete various downstream tasks effectively through appropriately set prompts. To meet the needs of our curriculum design, we have devised prompts for assessing the difficulty of correcting GEC training data, as indicated by the italicized text in Figure~\ref{fig:cl_method}. Assuming the training dataset is denoted as $D(S, T)$, the LLM scores the correction difficulty of the source-side sentences on a 10-point scale: 
\begin{equation}
p(Scores| [Prompt:S_{input}]; \theta).
\end{equation}
The $Scores$ may include natural language descriptions of the ratings. For specific examples, please refer to Appendix \ref{appendix:examples}.
In this scale, data with scores ranging from 1 to 3 are considered easy samples, 4 to 7 are medium difficulty samples and 8 to 10 are hard samples. Once the large language model has completed scoring the entire training set, we can obtain three curriculum datasets categorized by difficulty level: $D_{e}$, $D_{m}$, and $D_{h}$.

\subsection{LLM-based CL} 
When we construct curriculums of varying difficulty levels using an LLM, we adhere to the principles of curriculum learning, gradually progressing from easy to hard through iterative training. 

\paragraph{Easy-stage} During the initial stage, we only utilize easy samples for supervised fine-tuning until the model converges.
\begin{equation}
 \underset{\theta_{e}}{\arg\max} \left \{\sum_{\mathcal{D}_{\mathrm{e} }}\log P\left (t|s; \theta  \right )  \right \},
\end{equation}
where $\theta$ is based on the LLM initialization.

\paragraph{Medium-stage} During the second stage of learning intermediate difficulty curriculums, we steadily introduce new levels of difficulty data into the model trained during the first stage. Concurrently, to safeguard against the model forgetting the samples learned in the previous stage \cite{10.1007/978-981-19-7960-6_3}, we integrate the data from the preceding stage into the training process.
\begin{equation}
 \underset{\theta_{m}}{\arg\max} \left  \{ \sum_{\mathcal{D}_{\mathrm{e}} \cup \mathcal{D}_{\mathrm{m}}}\log P\left (t|s; \theta_{e}  \right )  \right \},
\end{equation}
where $\theta_{e}$ is based on the model initialization from the easy stage.

\paragraph{Hard-stage} Following a methodology similar to our previous training process, we persistently iterate by introducing increasingly challenging samples while reinforcing previously learned ones to safeguard against the model forgetting earlier knowledge \cite{10.1007/978-981-19-7960-6_3}. The training process concludes once the model ultimately converges, marking the end of the training.
\begin{equation}
 \underset{\theta_{h}}{\arg\max} \left  \{ \sum_{\mathcal{D}_{\mathrm{e}} \cup \mathcal{D}_{\mathrm{m}}\cup \mathcal{D}_{\mathrm{h}}}\log P\left (t|s; \theta_{m}  \right )  \right \},
\end{equation}
where $\theta_{h}$ is based on the model initialization from the medium stage, and $\theta_{m}$ is the parameter that encapsulates the entirety of the LLM-based CL training process.

\section{Experimental Settings and Results}
\subsection{Data and Evaluation}

For fine-tuning baseline PLMs, we adopt the approach of \citet{rothe-etal-2021-simple} by exclusively utilizing the English cLang8 data for fine-tuning. This decision was informed by their observation that additional fine-tuning on high-quality English datasets, such as FCE v2.1 \cite{yannakoudakis-etal-2011-new} and W\&I \cite{yannakoudakis2018developing}, resulted in decreased performance. It should be highlighted that deduplication has been executed on the cLang8 dataset.
In the selection process of easy, medium, and hard data samples by LLM, we exclude those samples that are correct-correct pairs. We mandate the LLM to rank exclusively the samples that are error-correct pairs.
We utilize the CoNLL13 \citep{ng-etal-2013-conll} for our validation set, while the test set is based on the widely used \emph{official-2014.combined.m2} version of CoNLL14 \citep{ng-etal-2014-conll}. We also use the BEA19 \citep{bryant-etal-2019-bea} development data when testing the BEA19 English test sets. Table \ref{Tab:datasets} displays the statistics of the datasets utilized in our experiments.

In assessing our model's performance, we employ the M2 scorer \citep{dahlmeier-ng-2012-better} for the CoNLL14 English test. The BEA19 English test is evaluated using ERRANT \citep{bryant-etal-2017-automatic}.
The $T$-test method is used to test the significance of our findings, with the exception of the BEA19 English test, as it constitutes a blind data.

\begin{table}[!ht]
\small
\centering
\resizebox{.45\textwidth}{!}{
\begin{tabular}{l|c|c|c|c|c}
\toprule
 \multicolumn{1}{c|}{\textsc{\textbf{Data}}}
& \multicolumn{1}{c|}{\textsc{\textbf{All}}} 
& \multicolumn{1}{c|}{\textsc{\textbf{Same}}} 
& \multicolumn{1}{c|}{\textsc{\textbf{Easy}}}
& \multicolumn{1}{c|}{\textsc{\textbf{Med.}}} 
& \multicolumn{1}{c}{\textsc{\textbf{Hard}}}\\
\cmidrule(lr){1-6}
CL8-train & 2.2M & 0.89M & 0.50M & 0.46M & 0.35M \\
\cmidrule(lr){1-6}
CoNLL13-test & 1,379 & - & - & - & - \\
BEA19-dev & 4,384  & - & -  &- & - \\
\cmidrule(lr){1-6}
CoNLL14-test & 1,312 & - & - & - & - \\
BEA19-test & 4,477 & - & - & - & - \\
\bottomrule
\end{tabular}}
\linespread{1}
\caption{Statistics of  the datasets utilized in our experiments. 
}
\label{Tab:datasets}
\vskip -1em
\end{table}

\subsection{Implementation Details and Training}
To optimally utilize the capabilities of large models to rank the difficulty levels of the GEC training set data, we choose the large-scale language model, LLaMA2-70b \cite{touvron2023llama2}, as a dedicated expert in determining grammatical difficulty. To enhance its ability to recognize and correct the difficulty level of grammatical errors, We carefully designed a prompt for it. In Appendix~\ref{appendix:examples}, we present the prompts we provided and the scoring results generated by LLaMA2-70b.

The main experiments are based on the T5-large and -xl series \cite{raffel2020exploring} pre-trained models and LLaMA2-7b and -13b series large-scale pre-trained language models \cite{touvron2023llama,touvron2023llama2}. In the course of our experiments, we use the \texttt{Huggingface}\footnote{\url{https://github.com/huggingface/transformers}} library to implement T5-large and -xl pre-trained models. We keep the implementation details and parameter settings consistent with \citet{fang-etal-2023-transgec}.
Additionally, we utilize the \texttt{Megtron-LM} \cite{shoeybi2020megatronlm} toolkit to implement LLaMA2-7b, -13b, and -70b series of LLMs for both SFT and inference. The full-parameter supervised fine-tuning was executed on 8 A800 GPUs. Further details regarding the training settings can be found in the Appendix~\ref{appendix:parameters}.

\begin{table*}[!ht]
\small
\centering
\resizebox{0.88\textwidth}{!}{
\begin{tabular}{l|rrr|rrr|rrr}
\toprule
{\multirow{2}{*}{\textsc{\textbf{System}}}}
& \multicolumn{3}{c|}{\textsc{\textbf{CoNLL14}}} 
& \multicolumn{3}{c|}{\textsc{\textbf{BEA19 (test)}}} 
& \multicolumn{3}{c}{\textsc{\textbf{BEA19 (dev)}}} \\
\cmidrule(r){2-10}
& {Pre.}  & {Rec.}  & \textsc{F$_{0.5}$}   & {Pre.}  & {Rec.}  & \textsc{F$_{0.5}$}  & {Pre.}  & {Rec.}  & \textsc{F$_{0.5}$} \\
\cmidrule{1-10}
\multicolumn{10}{c}{\textbf{Existing Baselines}} \\
\cmidrule{1-10}
GECToR \citep{omelianchuk-etal-2020-gector} & 77.5 & 40.1 & 65.3  & 79.2 & 53.9 & 72.4 & 66.0 & 33.8 & 55.5 \\
TagGEC \citep{stahlberg-kumar-2021-synthetic} & 72.8 & 49.5 & 66.6  & 72.1 & 64.4 & 70.4 & 59.5 & 41.3 & 54.7 \\
SADGEC \citep{sun-etal-2021-instantaneous} & 71.0 & 52.8 & 66.4  & - & - & 72.9 & - & - & - \\
T5-large \citep{rothe-etal-2021-simple} & - & - & 66.0 & - & - & 72.1 & - & - & - \\
T5-xl \citep{rothe-etal-2021-simple} & - & - & 67.7  & - & - & 73.9 & - & - & - \\
T5-xxl \citep{rothe-etal-2021-simple} & - & - & 68.8 & - & - & \bf 75.9 & - & - & - \\
TMTC \citep{lai-etal-2022-type} & 77.8 & 41.8 & 66.4  & 81.3 & 51.6 & 72.9 & 54.9 & 35.3 & 49.4 \\
TemplateGEC \citep{li-etal-2023-templategec} & 74.8 & 50.0 & 68.2  & 76.8 & 64.8 & 74.1 & 61.0 & 41.0 & 55.6 \\ %
MultimodalGEC \citep{fang-etal-2023-improving} & 73.6 & 52.7 & 68.2 & 75.5 & 67.9 & 73.9 & 61.0 & 45.3 & 57.0\\
MultiTaskGEC \citep{bout-etal-2023-efficient} & 75.4 & 51.2 & 68.9 & 78.2 & 65.5 & 75.3 & 68.6 & 46.5 & \bf 62.7 \\
ChatGPT (0-shot CoT) \citep{fang2023chatgpt} & 50.2 & 59.0 & 51.7 & 32.1 & 70.5 & 36.1 & - &- &-   \\
ChatGPT (3-shot CoT) \citep{fang2023chatgpt} & 51.3 & 62.4 & 53.2 & 34.0 & 70.2 & 37.9 & - & -&- \\
\cmidrule{1-10}  %
\multicolumn{10}{c}{\textbf{Our Implemented Baselines}} \\
\cmidrule{1-10}
T5-large GEC & 72.2 & 51.4 & 66.8 & 73.4 & 67.0 & 72.0 & 60.5 & 43.1 & 56.0 \\
T5-xl GEC & 73.7 & 52.4 & 68.1 & 75.9 & 69.4 & 74.5 & 61.9 & 45.0 & 57.6 \\
LLaMA2-7b GEC & 73.1 & 51.9 & 67.6 & 73.8 & 67.2 & 72.4 & 60.4 & 42.2& 55.6 \\
LLaMA2-13b GEC & 73.9 & 52.7 & 68.4 & 74.5 & 68.8 &73.3 & 62.2 & 43.7 & 57.3 \\
\cdashline{1-10}[2pt/2.5pt]\noalign{\vskip 0.5ex}
T5-large Len-based CL & 73.1 & 52.2 & 67.7 & 75.6 & 66.2 & 73.5 & 60.9 & 43.5 & 56.2 \\
T5-xl Len-based CL & 73.9 & 53.1 & 68.5 & 75.8 & 69.8 & 74.6 & 62.0 & 45.4 & 57.8 \\
LLaMA2-7b Len-based CL & 73.5 & 52.8 & 68.1 & 74.1 & 66.6 & 72.5 & 60.0 & 43.8 & 55.8 \\
LLaMA2-13b Len-based CL & 74.0 & 54.7 & 69.1 & 74.7 & 69.0 & 73.5 & 61.0 & 46.5 & 57.4 \\
\cmidrule{1-10}  %
\multicolumn{10}{c}{\textbf{Our Proposed Method}} \\
\cmidrule{1-10} %
\textbf{T5-large LLM-based CL} & 73.9 & 52.9 & 68.5$^{\dagger}$ & 76.0 & 66.8 & 74.0 & 60.7 & 44.1 & 56.5$^{\dagger}$ \\


\textbf{T5-xl LLM-based CL} & 74.6 & 	53.9 & 69.3$^{\dagger}$ & 77.9 & 67.7 & \bf 75.6 & 62.9 & 44.5 & \bf 58.1$^{\dagger}$ \\

\textbf{LLaMA2-7b LLM-based CL} & 73.6 & 53.4 & 68.4$^{\dagger}$ & 75.5 & 66.4 & 73.5 & 59.8 & 44.8 & 56.1$^{\dagger}$ \\

\textbf{LLaMA2-13b LLM-based CL} & 74.4 & 55.2 & 
\bf 69.6$^{\dagger}$ & 76.5 & 66.7 & 74.3 & 61.3 & 46.7 & 57.7$^{\dagger}$ \\
\bottomrule
\end{tabular}} 
\linespread{1}
\caption{Results on the CoNLL14 test and BEA19 test and development sets. T5-\textbf{xx}/ LLaMA2-\textbf{xx} GEC models are the implemented baselines which are fine-tuned on the same CLang8 English data as T5-large/xl \citep{rothe-etal-2021-simple}. The T5-\textbf{xx}/ LLaMA2-\textbf{xx} Len-based CL models refer to the difficulty in sorting based on the length of the training sentences, which is also the first time we have implemented this as comparable baselines. 
\textbf{Bold} values indicate the best F$_{0.5}$ scores across different systems. Statistically significant improvements over the T5-\textbf{xx}/ LLaMA2-\textbf{xx} GEC and \textbf{Len-based CL} baselines for the same model size, as validated using the $P${\_}value, with a level of $^{\dagger}p<0.01$.
}
\label{Tab:llm-based-CL_results}
\end{table*}
\subsection{Baselines}
To investigate the impact of our proposed LLM-based CL approach for GEC, our models are compared with the following baselines. 
The \textbf{GECToR} \cite{omelianchuk-etal-2020-gector} and \textbf{TMTC} \citep{lai-etal-2022-type} models employ a sequence tagging approach to enhance GEC performance through multi-stage training. 
The \textbf{TagGEC} \citep{stahlberg-kumar-2021-synthetic} model enhances the performance of GEC through data augmentation, specifically by generating synthetic data guided by error type tags. 
The GEC models, \textbf{SADGEC} \citep{sun-etal-2021-instantaneous}, \textbf{T5-large/xl/xxl} \cite{rothe-etal-2021-simple} are based on pre-trained language models.
SADGEC utilizes the BART pre-trained model, while T5-large/xl/xxl model uses T5 \cite{raffel2020exploring} as its backbone structure and is fine-tuned on the corresponding distilled cLang8 data, making it suitable for GEC tasks in various languages. 
The \textbf{TemplateGEC} \citep{li-etal-2023-templategec} model combines the seq2edit and seq2seq models to create a novel two-stage framework for identifying and rectifying errors. 
The \textbf{MultimodalGEC} \citep{fang-etal-2023-improving} model integrates speech modality information into the GEC task. 
The \textbf{MultitaskGEC} \citep{bout-etal-2023-efficient} suggested a multi-task pre-training approach and an optimization technique that substantially enhanced the efficiency of the GEC model. The results of \textbf{ChatGPT (0/3-shot CoT)} are the zero-shot and few-shot COT methods, which are reported by \citep{fang2023chatgpt}. 
In addition to the aforementioned strong existing baselines, we have also implemented a few new baselines for comparison. 
The \textbf{T5-large/xl GEC} and \textbf{LLaMA2-7/13b GEC} models, which are our implemented baselines, have been fine-tuned using the same cLang8 English data as the T5-large/xl \citep{rothe-etal-2021-simple}. 
The \textbf{T5-large/xl Len-based} and \textbf{LLaMA2-7/13b Len-based CL} models refer to the complexity of ranking based on the length of the training sentences. This is the first time we have implemented these as comparable baselines.

\subsection{Experimental Results}
To effectively illustrate the efficiency of our proposed LLM-based CL GEC models, we executed comparative experiments in conjunction with existing studies. The results of these experiments are clearly outlined in Table \ref{Tab:llm-based-CL_results}. 
As expected, the methods of learning through a designed curriculum and progressively training on the data, namely the Len-based CL and proposed LLM-based CL methods, have significantly better results than the different types of baseline models fine-tuned with the full mixed data ( i.e. T5-\textbf{xx}/ LLaMA2-\textbf{xx} GEC \textbf{\textsc{VS}} T5-\textbf{xx}/ LLaMA2-\textbf{xx} Len/ LLM-based CL). 
This has been demonstrated across three evaluation data sets: the CoNLL14 test, BEA19 test, and BEA19 development sets.
This marks our inaugural, comprehensive demonstration that curriculum learning proves effective in handling GEC tasks. 
It is noteworthy that, while the Len-based CL methods exhibit an enhanced performance compared to the implemented robust baseline models, the degree of their improvement is not as significant as that of our proposed LLM-based CL methods, when evaluated with the F$_{0.5}$ score under the same model size ( i.e. T5-\textbf{xx}/ LLaMA2-\textbf{xx} Len-based CL \textbf{\textsc{VS}} T5-\textbf{xx}/ LLaMA2-\textbf{xx} LLM-based CL).
This suggests that our introduced LLM-based CL methods demonstrate greater efficacy in utilizing the curriculum learning strategy for GEC task.

Interestingly, we observed that LLMs with a decode-only architecture, such as LLaMA2-7/13b, did not perform as well as models with encoder-decoder architecture, like T5-large/xl. Their parameters are significantly less than LLaMA2, with T5-large (770M) being much smaller than LLaMA2-7b, and T5-xl (3b) being substantially less than LLaMA2-13b. This was particularly evident in the performance on BEA19-test/dev, and the discrepancy was almost similar on CoNLL14 test. 
We speculate that this discrepancy may be due to the architectural differences having a significant impact on the GEC task. The GEC task requires the generation of correct source words for copying, making the encoder-decoder structure more suitable.
Additionally, as suggested by \citet{zhang2023multitask}, LLaMA series of LLMs also have hallucination issues that limit their performance on the GEC task.
While this issue warrants further investigation in future research, it is important to note that our proposed LLM-based CL method remains remarkably effective on models of the same series and size.
Appendix \ref{appendix:CL_pre_ds} displays the CL performance at various stages.

\section{Analysis}
\subsection{New Baselines Comparison}
During the training phase of curriculum learning, the common approach is to start with simple data and gradually introduce more complex samples for training. 
This prevents knowledge forgetting that may occur when encountering slightly more challenging data right after completing the introductory phase of the course. 
However, this approach might be contentious. For instance, we usually compare this method to a baseline that uses the entire default training data. 
However, during the process of curriculum learning, data from earlier stages might be repeatedly studied, which could lead to questions about the fairness of the baseline comparison. 
To address this issue, we designed an ablation study using the data of the same type and quantity for training. 
Table~\ref{Tab:newbaselines} presents the experimental results of the Len-based and LLM-based CL methods, utilizing T5-xl and LLaMA2-13b on the CoNLL14 and BEA19 test sets. 
The results clearly indicate that the newly established baselines indeed outperform conventional models trained solely with default data, yet they fail to reach the performance levels of the Len/ LLM-based CL methods. 
This observation underscores the implication that, even when accounting for data repetition learning factors, CL strategies can still markedly boost performance. 
Furthermore, it's noteworthy that the performance achieved through the LLM-based methods significantly surpasses that of the Len-based methods within the new baselines. 
This suggests that the selection of data via the LLM-based approach yields more beneficial results in model training. 

\begin{table}[!t]
\small
\centering
\resizebox{0.48\textwidth}{!}{
\begin{tabular}{l|c|c}
\toprule
{\multirow{2}{*}{\textsc{\textbf{System}}}}
& \multicolumn{1}{c|}{\textsc{\textbf{CoNLL14}}}
& \multicolumn{1}{c}{\textsc{\textbf{BEA19 (test)}}} \\
\cmidrule(r){2-3}
& \textsc{F$_{0.5}$} & \textsc{F$_{0.5}$} \\
\cmidrule(lr){1-3}
\multicolumn{3}{c}{\textbf{Based on T5-xl}} \\
\cmidrule(lr){1-3} %
Baseline (Default Data) & 68.1   & 74.5 \\
Len-based (3E+2M+1H) & 68.3  & 74.5 \\
LLM-based (3E+2M+1H)  & 68.6  & 74.8 \\ 
Len-based CL (E->EM->EMH)  &  68.5 & 74.6 \\ 
LLM-based CL (E->EM->EMH)  & \bf 69.3 &  \bf 75.6\\ 

\cmidrule(lr){1-3}
\multicolumn{3}{c}{\textbf{Based on LLaMA2-13b}} \\
\cmidrule(lr){1-3} %
Baseline (Default Data) & 68.4   & 73.3 \\
Len-based (3E+2M+1H) & 68.5  & 73.5 \\
LLM-based (3E+2M+1H)  & 68.8  & 73.8 \\ 
Len-based CL (E->EM->EMH)  & 69.1 & 73.5 \\ 
LLM-based CL (E->EM->EMH)  & \bf 69.6 &  \bf 74.3 \\ 
\bottomrule
\end{tabular}} 
\linespread{1}
\caption{The F$_{0.5}$ scores of both Len-based and LLM-based methods, which are trained on an equivalent amount of data using T5-xl and LLaMA2-13b, are evaluated on the CoNLL14 and BEA19 sets.
3E+2M+1H represents new baselines trained with a mix of 3 times easy, 2 times medium, and 1 time hard data. E->EM->EMH refers to the process of CL training. \textbf{Bold} values indicate the best scores across different models. 
}
\label{Tab:newbaselines}
\end{table}

\begin{table}[!ht]
\small
\centering
\resizebox{0.42\textwidth}{!}{
\begin{tabular}{l|ccc}
\toprule
{\multirow{2}{*}{\textsc{\textbf{\makecell[l]{\\System}}}}}
& \multicolumn{3}{c}{\textsc{\textbf{CoNLL14 (test)}}} \\
\cmidrule(r){2-4}
& Pre. & Rec. & F$_{0.5}$ \\
\cmidrule(lr){1-4}
\multicolumn{3}{c}{\textbf{Hard->Easy (Based on T5-large)}} \\
\cmidrule(lr){1-4} %
T5-large GEC & 72.2  & 51.4 & 66.8 \\
LLM-based Hard & 72.8  & 45.7 & 65.1 \\
~~~~~~~~~~~~~~~~~~~~~~+ Mid.  & 72.2  & 47.2 & 65.3 \\ 
~~~~~~~~~~~~~~~~~~~~~~~~~~+ Easy &  72.1 & 51.9 & 66.9 \\ 

\cmidrule(lr){1-4}
\multicolumn{3}{c}{\textbf{Easy->Hard (Based on T5-large)}} \\
\cmidrule(lr){1-4} %
LLM-based Easy & 75.9  & 36.6 & 62.5 \\
~~~~~~~~~~~~~~~~~~~~~~+ Mid.  & 73.9  & 44.8 & 65.4 \\ 
~~~~~~~~~~~~~~~~~~~~~~~~~~+ Hard & 73.9 & 52.9 & \bf 68.5 \\ 

\bottomrule
\end{tabular}} 
\linespread{1}
\caption{The performance of Hard->Easy and Easy->Hard CL strategies on T5-large. \textbf{Bold} values indicate the best scores.
}
\label{Tab:hard2easy}
\end{table}

\subsection{Learning Guidelines: From Hard to Easy}
To validate whether the model can only be effective through progressive training from easy to hard samples, we adjust the learning strategy to start with hard data and gradually incorporate simpler data. 
Table~\ref{Tab:hard2easy} displays the results of the varying difficulty stages of LLM-based CL method validation on the CoNLL14 test set using the T5-large model.	
We can observe that the performance in the first training phase between easy and hard data is significantly different. 
The \textbf{F$_{0.5}$} and \textbf{Recall} values for the hard data training in the first stage are significantly higher than those in easy data training. 
However, the \textbf{Precision} is not as good as in the case of easy data training. 
This suggests that the complexity and variety of errors in the hard data make it easier for the model to detect them, but correcting these detected errors presents a significant challenge. Meanwhile, the simplicity and lack of diversity in the errors of the easy data weaken the model's ability to detect errors and generalize, resulting in a low \textbf{Recall} value. 
As the training progresses to the second stage with the continuous addition of medium-difficulty data, the performance of both strategies is comparable. 
However, when the third stage of training is introduced, the \textbf{Easy->Hard} training strategy significantly outperforms the \textbf{Hard->Easy} strategy. 
In fact, the performance of the \textbf{Hard->Easy} training process is only on par with the baseline results upon its completion. 
This indicates that in GEC, the principle of CL should start from training with easy samples. Contrary strategies fail to yield noticeable advantages.

\begin{figure*}[!t]
\centering
\includegraphics[width=0.88\textwidth, trim=0 0 0 0]{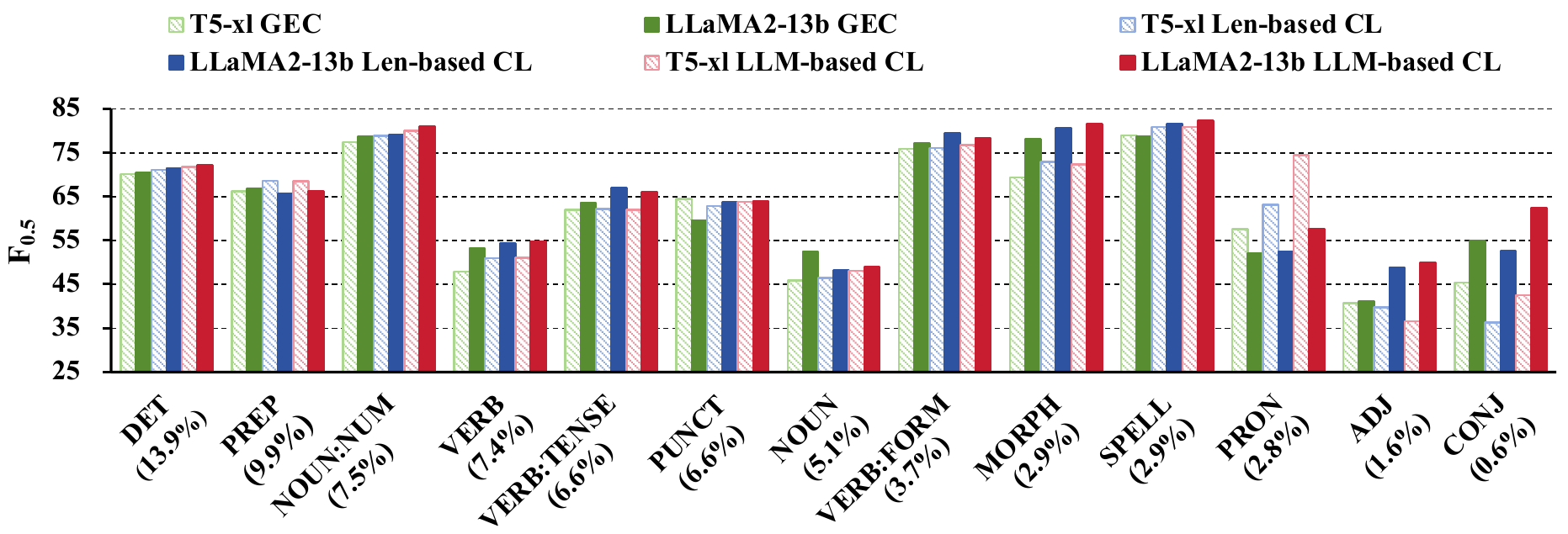}
\caption{
The F$_{0.5}$ scores reflect the performance on a selection of fine-grained error types within the CoNLL14 test set. 
The percentages provided in brackets represent the distribution of each error type. 
The findings indicate that our LLM-based CL method can significantly improve performance in correcting a majority of these fine-grained error types, surpassing the efficiency of both the Len-based method and other baseline approaches.
}
\label{fig:error_type_analy}
\end{figure*}

\subsection{Types of Error Analysis}
We employ the ERRANT toolkit \citep{bryant-etal-2017-automatic} to examine the performance of GEC systems in correcting diverse types of errors. 
Figure \ref{fig:error_type_analy} illustrates the performance of T5-xl and LLaMA2-13b's Baseline, Len-based CL, and LLM-based CL methods on the CoNLL14 test set in terms of POS-based fine-grained error types. 
It is evident from the figure that, regardless of whether it is the baseline or CL method, LLaMA2-13b outperforms the T5-xl model in the majority of fine-grained error types (i.e., DET, NOUN:NUM, VERB, VERB:TENSE, NOUN, VERB:FORM, ADJ, CONJ). 
However, there are a few error types where the trend is reversed, such as in the case of PREP and PRON performance. 
This could potentially be attributed to the differences in the structures of the two models.
Moreover, the LLM-based approach evidently shows superior performance over the Len-based CL in many error types, including but not limited to DET, NOUN:NUM, VERB, PUNCT, NOUN, MORPH, SPELL, PRON, and CONJ.
Overall, the CL methods exhibit good performance across various errors in GEC, particularly our proposed LLM-based CL method, which excels in error correction performance. 
In addition, we provide the Operation-Level error types for the BEA19 development set in Appendix \ref{appendix:error_type_ol}.


\section{Related Work}
\paragraph{GEC in Language Models} 
The rise of PLMs has led to a surge in research efforts that utilize these models to boost the efficiency of GEC. Recent studies show the positive impact of PLMs on the enhancement of GEC tasks. For example, a study conducted by \citet{choe-etal-2019-neural} used sequential transfer learning to adapt pre-trained Transformer models to the specific needs of GEC. Another strategy was implemented by \citet{kaneko-etal-2020-encoder}, who chose to initialize an encoder-decoder GEC model with pre-trained BERT weights, thereby improving GEC performance. Further, \citet{katsumata-komachi-2020-stronger} incorporated the pre-trained BART model as a universal pre-trained encoder-decoder model for GEC, while \citet{rothe-etal-2021-simple} opted for a pre-trained T5 model to distill the GEC corpus, incorporating this pre-trained framework into the network for distilled GEC training and achieving notable results. Apart from these methods, there are many other research efforts that use these pre-trained models as a strong baseline for further study \citep{wan-etal-2020-improving,tarnavskyi-etal-2022-ensembling,qorib-etal-2022-frustratingly,fang-etal-2023-improving,zhang-etal-2022-mucgec,fang-etal-2023-transgec}. Regarding recent studies on the GEC capabilities of large language models (LLMs), notable examples include those by \citet{fang2023chatgpt} and \citet{loem-etal-2023-exploring}. \citet{zhang2023multitask} utilize the instruct tuning of LLaMA for the task of GEC. Despite the aforementioned work discussing the use of LMs as a robust baseline, all data is traditionally incorporated for fine-tuning during the training process. There has been no discussion about the impact of data difficulty on language models or LLMs in the GEC task. 

\paragraph{Curriculum Learning} 
The concept of Curriculum Learning (CL) has been introduced in the field of machine learning, and is used for tasks such as image classification and language modeling \citep{10.1145/1553374.1553380}. This method entails a categorization of examples into easy and difficult and the organization of a curriculum from simple to complex. Studies have shown that during training, a model can take advantage of the CL strategy \citep{ren2019learning,data-parameters}. The crucial aspect here is to devise the right challenging examples during the curriculum design process.
To elaborate further, CL has played a pivotal role in NLP applications including but not limited to dependency parsing \citep{spitkovsky-etal-2010-baby}, sentiment analysis \citep{Cirik2016VisualizingAU}, neural machine translation (NMT) \citep{kocmi-bojar-2017-curriculum,kumar-etal-2019-reinforcement,Zhang2018AnEE,liu-etal-2020-norm,zhou-etal-2020-uncertainty} and reading comprehension \citep{tay-etal-2019-simple}. The key to effective CL lies in setting appropriate difficulty levels for different tasks. To achieve this goal, numerous strategies have been proposed, which are deemed to guide the step-by-step learning of the model. For instance, \citet{zhou-etal-2020-uncertainty} introduced an uncertainty-aware CL concept. They use the perplexity of training examples as a criterion for curriculum difficulty and employ a self-adaptive learning strategy for the NMT task. \citet{zhao-etal-2020-knowledge-grounded} suggest a method for generating responses by incorporating a knowledge selection module into GPT-2. \citet{10.1145/3510003.3510062} have incorporated CL to structure the transformed data from easy to complex. This is done to fine-tune pre-existing models that are pre-trained for the purpose of understanding source code. At present, our paper is the first to consider using CL methods in the GEC task. It is also the first time that a LLM is used to automatically analyze the difficulty level of data, and this method has achieved significant results. 

\section{Conclusion}
This paper proposes a novel method for CL based on LLMs, called the LLM-based CL method. Compared to conventional CL approaches
, our proposed LLM-based CL method aligns closely with curriculums designed by human experts, indicating that leveraging the powerful semantic understanding and discriminative capabilities of LLMs enables the differentiation and reasonable arrangement of GEC training data from easy to hard samples into different curriculums. Upon obtaining GEC data of varying difficulty curriculum levels, we iteratively train and refine models from easy to hard samples using the T5 and LLaMA series. Through testing and analysis on various English GEC benchmarks, 
we find that the proposed LLM-based CL method significantly outperforms baseline models and previous CL methods. Furthermore, we discuss the sequence of training in CL forms and find that models only benefit from learning processes that progress from easy to hard; conversely, learning from hard to easy is almost ineffective, validating the effectiveness of CL in GEC. In addition to comparing against the traditional baseline of training data directly at once, we also compare CL step-by-step iterative training with training by adding the same amount of data at once as new baselines, finding that progressively adding difficult samples enhances GEC model performance.

\section*{Limitations}
The paper aims to propose an LLM-based curriculum learning paradigm to enhancing the English GEC performance. While experiments and analyses validate the effectiveness of this method, there are still aome limitations: Firstly, the process of annotating and scoring millions of training data instances using a large-scale language model to differentiate between different courses is time-consuming; however, this may represent a trade-off between performance and time. Secondly, we have only validated the effectiveness of constructing a new curriculum learning paradigm using a large-scale language model in the context of the GEC task. Whether this approach is effective for other NLP tasks remains to be further explored. In the future, we plan to extend our research to include a broader range of NLP tasks.

\section*{Acknowledgements}
This work was supported in part by the Science and Technology Development Fund, Macau SAR (Grant No. FDCT/060/2022/AFJ, the mainland China collaboration project, National Natural Science Foundation of China Grant No. 62261160648), the Science and Technology Development Fund, Macau SAR (Grant No. FDCT/0070/2022/AMJ, the mainland China collaboration project, China Strategic Scientific and Technological Innovation Cooperation Project Grant No. 2022YFE0204900), the Multi-year Research Grant from the University of Macau (Grant No. MYRG-GRG2023-00006-FST-UMDF), and the Tencent AI Lab Rhino-Bird Gift Fund (Grant No. EF2023-00151-FST). This work was performed in part at SICC which is supported by SKL-IOTSC, and HPCC supported by ICTO of the University of Macau. We would like to thank the LLM team at Bilibili Inc. for their support with high-performance GPUs.


\bibliography{anthology_1,custom}

\begin{thebibliography}{55}
\providecommand{\natexlab}[1]{#1}

\bibitem[{Albalak et~al.(2024)Albalak, Elazar, Xie, Longpre, Lambert, Wang, Muennighoff, Hou, Pan, Jeong, Raffel, Chang, Hashimoto, and Wang}]{albalak2024survey}
Alon Albalak, Yanai Elazar, Sang~Michael Xie, Shayne Longpre, Nathan Lambert, Xinyi Wang, Niklas Muennighoff, Bairu Hou, Liangming Pan, Haewon Jeong, Colin Raffel, Shiyu Chang, Tatsunori Hashimoto, and William~Yang Wang. 2024.
\newblock \href {https://arxiv.org/abs/2402.16827} {A survey on data selection for language models}.
\newblock \emph{arXiv preprint arXiv:2402.16827}.

\bibitem[{Bengio et~al.(2009)Bengio, Louradour, Collobert, and Weston}]{10.1145/1553374.1553380}
Yoshua Bengio, J\'{e}r\^{o}me Louradour, Ronan Collobert, and Jason Weston. 2009.
\newblock \href {https://doi.org/10.1145/1553374.1553380} {Curriculum learning}.
\newblock In \emph{Proceedings of the 26th Annual International Conference on Machine Learning}, ICML '09, page 41–48, New York, NY, USA. Association for Computing Machinery.

\bibitem[{Bout et~al.(2023)Bout, Podolskiy, Nikolenko, and Piontkovskaya}]{bout-etal-2023-efficient}
Andrey Bout, Alexander Podolskiy, Sergey Nikolenko, and Irina Piontkovskaya. 2023.
\newblock \href {https://doi.org/10.18653/v1/2023.emnlp-main.355} {Efficient grammatical error correction via multi-task training and optimized training schedule}.
\newblock In \emph{Proceedings of the 2023 Conference on Empirical Methods in Natural Language Processing}, pages 5800--5816, Singapore. Association for Computational Linguistics.

\bibitem[{Brown et~al.(2020)Brown, Mann, Ryder, Subbiah, Kaplan, Dhariwal, Neelakantan, Shyam, Sastry, Askell, Agarwal, Herbert-Voss, Krueger, Henighan, Child, Ramesh, Ziegler, Wu, Winter, Hesse, Chen, Sigler, Litwin, Gray, Chess, Clark, Berner, McCandlish, Radford, Sutskever, and Amodei}]{NEURIPS2020_1457c0d6}
Tom Brown, Benjamin Mann, Nick Ryder, Melanie Subbiah, Jared~D Kaplan, Prafulla Dhariwal, Arvind Neelakantan, Pranav Shyam, Girish Sastry, Amanda Askell, Sandhini Agarwal, Ariel Herbert-Voss, Gretchen Krueger, Tom Henighan, Rewon Child, Aditya Ramesh, Daniel Ziegler, Jeffrey Wu, Clemens Winter, Chris Hesse, Mark Chen, Eric Sigler, Mateusz Litwin, Scott Gray, Benjamin Chess, Jack Clark, Christopher Berner, Sam McCandlish, Alec Radford, Ilya Sutskever, and Dario Amodei. 2020.
\newblock \href {https://proceedings.neurips.cc/paper_files/paper/2020/file/1457c0d6bfcb4967418bfb8ac142f64a-Paper.pdf} {Language models are few-shot learners}.
\newblock In \emph{Advances in Neural Information Processing Systems}, volume~33, pages 1877--1901. Curran Associates, Inc.

\bibitem[{Bryant et~al.(2019)Bryant, Felice, Andersen, and Briscoe}]{bryant-etal-2019-bea}
Christopher Bryant, Mariano Felice, {\O}istein~E. Andersen, and Ted Briscoe. 2019.
\newblock \href {https://doi.org/10.18653/v1/W19-4406} {The {BEA}-2019 shared task on grammatical error correction}.
\newblock In \emph{Proceedings of the Fourteenth Workshop on Innovative Use of NLP for Building Educational Applications}, pages 52--75, Florence, Italy. Association for Computational Linguistics.

\bibitem[{Bryant et~al.(2017)Bryant, Felice, and Briscoe}]{bryant-etal-2017-automatic}
Christopher Bryant, Mariano Felice, and Ted Briscoe. 2017.
\newblock \href {https://doi.org/10.18653/v1/P17-1074} {Automatic annotation and evaluation of error types for grammatical error correction}.
\newblock In \emph{Proceedings of the 55th Annual Meeting of the Association for Computational Linguistics (Volume 1: Long Papers)}, pages 793--805, Vancouver, Canada. Association for Computational Linguistics.

\bibitem[{Choe et~al.(2019)Choe, Ham, Park, and Yoon}]{choe-etal-2019-neural}
Yo~Joong Choe, Jiyeon Ham, Kyubyong Park, and Yeoil Yoon. 2019.
\newblock \href {https://doi.org/10.18653/v1/W19-4423} {A neural grammatical error correction system built on better pre-training and sequential transfer learning}.
\newblock In \emph{Proceedings of the Fourteenth Workshop on Innovative Use of NLP for Building Educational Applications}, pages 213--227, Florence, Italy. Association for Computational Linguistics.

\bibitem[{Cirik et~al.(2016)Cirik, Hovy, and Morency}]{Cirik2016VisualizingAU}
Volkan Cirik, Eduard~H. Hovy, and Louis-Philippe Morency. 2016.
\newblock \href {https://api.semanticscholar.org/CorpusID:17896684} {Visualizing and understanding curriculum learning for long short-term memory networks}.
\newblock \emph{ArXiv}, abs/1611.06204.

\bibitem[{Dahlmeier and Ng(2012)}]{dahlmeier-ng-2012-better}
Daniel Dahlmeier and Hwee~Tou Ng. 2012.
\newblock \href {https://aclanthology.org/N12-1067} {Better evaluation for grammatical error correction}.
\newblock In \emph{Proceedings of the 2012 Conference of the North {A}merican Chapter of the Association for Computational Linguistics: Human Language Technologies}, pages 568--572, Montr{\'e}al, Canada. Association for Computational Linguistics.

\bibitem[{Fang et~al.(2023{\natexlab{a}})Fang, Hu, Wong, Wan, Chao, and Chang}]{fang-etal-2023-improving}
Tao Fang, Jinpeng Hu, Derek~F. Wong, Xiang Wan, Lidia~S. Chao, and Tsung-Hui Chang. 2023{\natexlab{a}}.
\newblock \href {https://doi.org/10.18653/v1/2023.findings-acl.594} {Improving grammatical error correction with multimodal feature integration}.
\newblock In \emph{Findings of the Association for Computational Linguistics: ACL 2023}, pages 9328--9344, Toronto, Canada. Association for Computational Linguistics.

\bibitem[{Fang et~al.(2023{\natexlab{b}})Fang, Liu, Wong, Zhan, Ding, Chao, Tao, and Zhang}]{fang-etal-2023-transgec}
Tao Fang, Xuebo Liu, Derek~F. Wong, Runzhe Zhan, Liang Ding, Lidia~S. Chao, Dacheng Tao, and Min Zhang. 2023{\natexlab{b}}.
\newblock \href {https://doi.org/10.18653/v1/2023.findings-acl.223} {{T}rans{GEC}: Improving grammatical error correction with translationese}.
\newblock In \emph{Findings of the Association for Computational Linguistics: ACL 2023}, pages 3614--3633, Toronto, Canada. Association for Computational Linguistics.

\bibitem[{Fang et~al.(2023{\natexlab{c}})Fang, Yang, Lan, Wong, Hu, Chao, and Zhang}]{fang2023chatgpt}
Tao Fang, Shu Yang, Kaixin Lan, Derek~F. Wong, Jinpeng Hu, Lidia~S. Chao, and Yue Zhang. 2023{\natexlab{c}}.
\newblock \href {https://arxiv.org/abs/2304.01746} {Is chatgpt a highly fluent grammatical error correction system? a comprehensive evaluation}.
\newblock \emph{arXiv preprint arXiv:2304.01746}.

\bibitem[{Hendy et~al.(2023)Hendy, Abdelrehim, Sharaf, Raunak, Gabr, Matsushita, Kim, Afify, and Awadalla}]{hendy2023good}
Amr Hendy, Mohamed Abdelrehim, Amr Sharaf, Vikas Raunak, Mohamed Gabr, Hitokazu Matsushita, Young~Jin Kim, Mohamed Afify, and Hany~Hassan Awadalla. 2023.
\newblock \href {https://arxiv.org/pdf/2302.09210.pdf} {How good are gpt models at machine translation? a comprehensive evaluation}.
\newblock \emph{arXiv preprint arXiv:2302.09210}.

\bibitem[{Hui et~al.(2022)Hui, Feng, and Zhang}]{10.1007/978-981-19-7960-6_3}
Ziyang Hui, Chong Feng, and Tianfu Zhang. 2022.
\newblock Review-based curriculum learning for neural machine translation.
\newblock In \emph{Machine Translation}, pages 24--36, Singapore. Springer Nature Singapore.

\bibitem[{Jiao et~al.(2023)Jiao, Wang, Huang, Wang, and Tu}]{jiao2023chatgpt}
Wenxiang Jiao, Wenxuan Wang, Jen-tse Huang, Xing Wang, and Zhaopeng Tu. 2023.
\newblock \href {https://www.researchgate.net/publication/367359399_Is_ChatGPT_A_Good_Translator_A_Preliminary_Study} {Is chatgpt a good translator? a preliminary study}.
\newblock \emph{arXiv preprint arXiv:2301.08745}, 1(10).

\bibitem[{Kaneko et~al.(2020)Kaneko, Mita, Kiyono, Suzuki, and Inui}]{kaneko-etal-2020-encoder}
Masahiro Kaneko, Masato Mita, Shun Kiyono, Jun Suzuki, and Kentaro Inui. 2020.
\newblock \href {https://doi.org/10.18653/v1/2020.acl-main.391} {Encoder-decoder models can benefit from pre-trained masked language models in grammatical error correction}.
\newblock In \emph{Proceedings of the 58th Annual Meeting of the Association for Computational Linguistics}, pages 4248--4254, Online. Association for Computational Linguistics.

\bibitem[{Katsumata and Komachi(2020)}]{katsumata-komachi-2020-stronger}
Satoru Katsumata and Mamoru Komachi. 2020.
\newblock \href {https://aclanthology.org/2020.aacl-main.83} {Stronger baselines for grammatical error correction using a pretrained encoder-decoder model}.
\newblock In \emph{Proceedings of the 1st Conference of the Asia-Pacific Chapter of the Association for Computational Linguistics and the 10th International Joint Conference on Natural Language Processing}, pages 827--832, Suzhou, China. Association for Computational Linguistics.

\bibitem[{Kocmi and Bojar(2017)}]{kocmi-bojar-2017-curriculum}
Tom Kocmi and Ond{\v{r}}ej Bojar. 2017.
\newblock \href {https://doi.org/10.26615/978-954-452-049-6_050} {Curriculum learning and minibatch bucketing in neural machine translation}.
\newblock In \emph{Proceedings of the International Conference Recent Advances in Natural Language Processing, {RANLP} 2017}, pages 379--386, Varna, Bulgaria. INCOMA Ltd.

\bibitem[{Kumar et~al.(2019)Kumar, Foster, Cherry, and Krikun}]{kumar-etal-2019-reinforcement}
Gaurav Kumar, George Foster, Colin Cherry, and Maxim Krikun. 2019.
\newblock \href {https://doi.org/10.18653/v1/N19-1208} {Reinforcement learning based curriculum optimization for neural machine translation}.
\newblock In \emph{Proceedings of the 2019 Conference of the North {A}merican Chapter of the Association for Computational Linguistics: Human Language Technologies, Volume 1 (Long and Short Papers)}, pages 2054--2061, Minneapolis, Minnesota. Association for Computational Linguistics.

\bibitem[{Lai et~al.(2022)Lai, Zhou, Zeng, Li, Li, Cao, and Su}]{lai-etal-2022-type}
Shaopeng Lai, Qingyu Zhou, Jiali Zeng, Zhongli Li, Chao Li, Yunbo Cao, and Jinsong Su. 2022.
\newblock \href {https://doi.org/10.18653/v1/2022.findings-acl.254} {Type-driven multi-turn corrections for grammatical error correction}.
\newblock In \emph{Findings of the Association for Computational Linguistics: ACL 2022}, pages 3225--3236, Dublin, Ireland. Association for Computational Linguistics.

\bibitem[{Li et~al.(2023)Li, Liu, Wang, Gong, Wong, Gao, Huang, and Zhang}]{li-etal-2023-templategec}
Yinghao Li, Xuebo Liu, Shuo Wang, Peiyuan Gong, Derek~F. Wong, Yang Gao, Heyan Huang, and Min Zhang. 2023.
\newblock \href {https://doi.org/10.18653/v1/2023.acl-long.380} {{T}emplate{GEC}: Improving grammatical error correction with detection template}.
\newblock In \emph{Proceedings of the 61st Annual Meeting of the Association for Computational Linguistics (Volume 1: Long Papers)}, pages 6878--6892, Toronto, Canada. Association for Computational Linguistics.

\bibitem[{Liu et~al.(2020)Liu, Lai, Wong, and Chao}]{liu-etal-2020-norm}
Xuebo Liu, Houtim Lai, Derek~F. Wong, and Lidia~S. Chao. 2020.
\newblock \href {https://doi.org/10.18653/v1/2020.acl-main.41} {Norm-based curriculum learning for neural machine translation}.
\newblock In \emph{Proceedings of the 58th Annual Meeting of the Association for Computational Linguistics}, pages 427--436, Online. Association for Computational Linguistics.

\bibitem[{Loem et~al.(2023)Loem, Kaneko, Takase, and Okazaki}]{loem-etal-2023-exploring}
Mengsay Loem, Masahiro Kaneko, Sho Takase, and Naoaki Okazaki. 2023.
\newblock \href {https://doi.org/10.18653/v1/2023.bea-1.18} {Exploring effectiveness of {GPT}-3 in grammatical error correction: A study on performance and controllability in prompt-based methods}.
\newblock In \emph{Proceedings of the 18th Workshop on Innovative Use of NLP for Building Educational Applications (BEA 2023)}, pages 205--219, Toronto, Canada. Association for Computational Linguistics.

\bibitem[{Minaee et~al.(2024)Minaee, Mikolov, Nikzad, Chenaghlu, Socher, Amatriain, and Gao}]{minaee2024large}
Shervin Minaee, Tomas Mikolov, Narjes Nikzad, Meysam Chenaghlu, Richard Socher, Xavier Amatriain, and Jianfeng Gao. 2024.
\newblock \href {https://arxiv.org/abs/2402.06196} {Large language models: A survey}.
\newblock \emph{arXiv preprint arXiv:2402.06196}.

\bibitem[{Ng et~al.(2014)Ng, Wu, Briscoe, Hadiwinoto, Susanto, and Bryant}]{ng-etal-2014-conll}
Hwee~Tou Ng, Siew~Mei Wu, Ted Briscoe, Christian Hadiwinoto, Raymond~Hendy Susanto, and Christopher Bryant. 2014.
\newblock \href {https://doi.org/10.3115/v1/W14-1701} {The {C}o{NLL}-2014 shared task on grammatical error correction}.
\newblock In \emph{Proceedings of the Eighteenth Conference on Computational Natural Language Learning: Shared Task}, pages 1--14, Baltimore, Maryland. Association for Computational Linguistics.

\bibitem[{Ng et~al.(2013)Ng, Wu, Wu, Hadiwinoto, and Tetreault}]{ng-etal-2013-conll}
Hwee~Tou Ng, Siew~Mei Wu, Yuanbin Wu, Christian Hadiwinoto, and Joel Tetreault. 2013.
\newblock \href {https://aclanthology.org/W13-3601} {The {C}o{NLL}-2013 shared task on grammatical error correction}.
\newblock In \emph{Proceedings of the Seventeenth Conference on Computational Natural Language Learning: Shared Task}, pages 1--12, Sofia, Bulgaria. Association for Computational Linguistics.

\bibitem[{Omelianchuk et~al.(2020)Omelianchuk, Atrasevych, Chernodub, and Skurzhanskyi}]{omelianchuk-etal-2020-gector}
Kostiantyn Omelianchuk, Vitaliy Atrasevych, Artem Chernodub, and Oleksandr Skurzhanskyi. 2020.
\newblock \href {https://doi.org/10.18653/v1/2020.bea-1.16} {{GECT}o{R} {--} grammatical error correction: Tag, not rewrite}.
\newblock In \emph{Proceedings of the Fifteenth Workshop on Innovative Use of NLP for Building Educational Applications}, pages 163--170, Seattle, WA, USA → Online. Association for Computational Linguistics.

\bibitem[{OpenAI(2024)}]{openai2024gpt4}
OpenAI. 2024.
\newblock \href {https://arxiv.org/abs/2303.08774} {Gpt-4 technical report}.
\newblock \emph{arXiv preprint arXiv:2303.08774}.

\bibitem[{Pan et~al.(2023)Pan, Chen, Xu, Che, and Qin}]{pan2023preliminary}
Wenbo Pan, Qiguang Chen, Xiao Xu, Wanxiang Che, and Libo Qin. 2023.
\newblock \href {https://arxiv.org/abs/2304.04256} {A preliminary evaluation of chatgpt for zero-shot dialogue understanding}.
\newblock \emph{arXiv preprint arXiv:2304.04256}.

\bibitem[{Platanios et~al.(2019)Platanios, Stretcu, Neubig, Poczos, and Mitchell}]{platanios-etal-2019-competence}
Emmanouil~Antonios Platanios, Otilia Stretcu, Graham Neubig, Barnabas Poczos, and Tom Mitchell. 2019.
\newblock \href {https://doi.org/10.18653/v1/N19-1119} {Competence-based curriculum learning for neural machine translation}.
\newblock In \emph{Proceedings of the 2019 Conference of the North {A}merican Chapter of the Association for Computational Linguistics: Human Language Technologies, Volume 1 (Long and Short Papers)}, pages 1162--1172, Minneapolis, Minnesota. Association for Computational Linguistics.

\bibitem[{Qorib et~al.(2022)Qorib, Na, and Ng}]{qorib-etal-2022-frustratingly}
Muhammad~Reza Qorib, Seung-Hoon Na, and Hwee~Tou Ng. 2022.
\newblock \href {https://doi.org/10.18653/v1/2022.naacl-main.143} {Frustratingly easy system combination for grammatical error correction}.
\newblock In \emph{Proceedings of the 2022 Conference of the North American Chapter of the Association for Computational Linguistics: Human Language Technologies}, pages 1964--1974, Seattle, United States. Association for Computational Linguistics.

\bibitem[{Raffel et~al.(2020)Raffel, Shazeer, Roberts, Lee, Narang, Matena, Zhou, Li, Liu et~al.}]{raffel2020exploring}
Colin Raffel, Noam Shazeer, Adam Roberts, Katherine Lee, Sharan Narang, Michael Matena, Yanqi Zhou, Wei Li, Peter~J Liu, et~al. 2020.
\newblock \href {https://jmlr.org/papers/volume21/20-074/20-074.pdf} {Exploring the limits of transfer learning with a unified text-to-text transformer.}
\newblock \emph{Journal of Machine Learning Research}, 21(140):1--67.

\bibitem[{Ren et~al.(2019)Ren, Zeng, Yang, and Urtasun}]{ren2019learning}
Mengye Ren, Wenyuan Zeng, Bin Yang, and Raquel Urtasun. 2019.
\newblock \href {https://arxiv.org/abs/1803.09050} {Learning to reweight examples for robust deep learning}.
\newblock \emph{arXiv.1803.09050}.

\bibitem[{Rothe et~al.(2021)Rothe, Mallinson, Malmi, Krause, and Severyn}]{rothe-etal-2021-simple}
Sascha Rothe, Jonathan Mallinson, Eric Malmi, Sebastian Krause, and Aliaksei Severyn. 2021.
\newblock \href {https://doi.org/10.18653/v1/2021.acl-short.89} {A simple recipe for multilingual grammatical error correction}.
\newblock In \emph{Proceedings of the 59th Annual Meeting of the Association for Computational Linguistics and the 11th International Joint Conference on Natural Language Processing (Volume 2: Short Papers)}, pages 702--707, Online. Association for Computational Linguistics.

\bibitem[{Saxena et~al.(2019)Saxena, Tuzel, and DeCoste}]{data-parameters}
Shreyas Saxena, Oncel Tuzel, and Dennis DeCoste. 2019.
\newblock \href {https://papers.nips.cc/paper/9289-data-parameters-a-new-family-of-parameters-for-learning-a-differentiable-curriculum.pdf} {Data parameters: A new family of parameters for learning a differentiable curriculum}.
\newblock In \emph{NeurIPS}.

\bibitem[{Shoeybi et~al.(2020)Shoeybi, Patwary, Puri, LeGresley, Casper, and Catanzaro}]{shoeybi2020megatronlm}
Mohammad Shoeybi, Mostofa Patwary, Raul Puri, Patrick LeGresley, Jared Casper, and Bryan Catanzaro. 2020.
\newblock \href {https://arxiv.org/abs/1909.08053} {Megatron-lm: Training multi-billion parameter language models using model parallelism}.
\newblock \emph{Preprint}, arXiv:1909.08053.

\bibitem[{Spitkovsky et~al.(2010)Spitkovsky, Alshawi, and Jurafsky}]{spitkovsky-etal-2010-baby}
Valentin~I. Spitkovsky, Hiyan Alshawi, and Daniel Jurafsky. 2010.
\newblock \href {https://aclanthology.org/N10-1116} {From baby steps to leapfrog: How {``}less is more{''} in unsupervised dependency parsing}.
\newblock In \emph{Human Language Technologies: The 2010 Annual Conference of the North {A}merican Chapter of the Association for Computational Linguistics}, pages 751--759, Los Angeles, California. Association for Computational Linguistics.

\bibitem[{Stahlberg and Kumar(2021)}]{stahlberg-kumar-2021-synthetic}
Felix Stahlberg and Shankar Kumar. 2021.
\newblock \href {https://aclanthology.org/2021.bea-1.4} {Synthetic data generation for grammatical error correction with tagged corruption models}.
\newblock In \emph{Proceedings of the 16th Workshop on Innovative Use of NLP for Building Educational Applications}, pages 37--47, Online. Association for Computational Linguistics.

\bibitem[{Sun et~al.(2021)Sun, Ge, Wei, and Wang}]{sun-etal-2021-instantaneous}
Xin Sun, Tao Ge, Furu Wei, and Houfeng Wang. 2021.
\newblock \href {https://doi.org/10.18653/v1/2021.acl-long.462} {Instantaneous grammatical error correction with shallow aggressive decoding}.
\newblock In \emph{Proceedings of the 59th Annual Meeting of the Association for Computational Linguistics and the 11th International Joint Conference on Natural Language Processing (Volume 1: Long Papers)}, pages 5937--5947, Online. Association for Computational Linguistics.

\bibitem[{Tarnavskyi et~al.(2022)Tarnavskyi, Chernodub, and Omelianchuk}]{tarnavskyi-etal-2022-ensembling}
Maksym Tarnavskyi, Artem Chernodub, and Kostiantyn Omelianchuk. 2022.
\newblock \href {https://doi.org/10.18653/v1/2022.acl-long.266} {Ensembling and knowledge distilling of large sequence taggers for grammatical error correction}.
\newblock In \emph{Proceedings of the 60th Annual Meeting of the Association for Computational Linguistics (Volume 1: Long Papers)}, pages 3842--3852, Dublin, Ireland. Association for Computational Linguistics.

\bibitem[{Tay et~al.(2019)Tay, Wang, Luu, Fu, Phan, Yuan, Rao, Hui, and Zhang}]{tay-etal-2019-simple}
Yi~Tay, Shuohang Wang, Anh~Tuan Luu, Jie Fu, Minh~C. Phan, Xingdi Yuan, Jinfeng Rao, Siu~Cheung Hui, and Aston Zhang. 2019.
\newblock \href {https://doi.org/10.18653/v1/P19-1486} {Simple and effective curriculum pointer-generator networks for reading comprehension over long narratives}.
\newblock In \emph{Proceedings of the 57th Annual Meeting of the Association for Computational Linguistics}, pages 4922--4931, Florence, Italy. Association for Computational Linguistics.

\bibitem[{Touvron et~al.(2023{\natexlab{a}})Touvron, Lavril, Izacard, Martinet, Lachaux, Lacroix, Rozi{\`e}re, Goyal, Hambro, Azhar et~al.}]{touvron2023llama}
Hugo Touvron, Thibaut Lavril, Gautier Izacard, Xavier Martinet, Marie-Anne Lachaux, Timoth{\'e}e Lacroix, Baptiste Rozi{\`e}re, Naman Goyal, Eric Hambro, Faisal Azhar, et~al. 2023{\natexlab{a}}.
\newblock \href {https://arxiv.org/abs/2302.13971} {Llama: Open and efficient foundation language models}.
\newblock \emph{arXiv preprint arXiv:2302.13971}.

\bibitem[{Touvron et~al.(2023{\natexlab{b}})Touvron, Martin, Stone, Albert, Almahairi, Babaei, Bashlykov, Batra, Bhargava, Bhosale, Bikel, Blecher, Ferrer, Chen, Cucurull, Esiobu, Fernandes, Fu, Fu, Fuller, Gao, Goswami, Goyal, Hartshorn, Hosseini, Hou, Inan, Kardas, Kerkez, Khabsa, Kloumann, Korenev, Koura, Lachaux, Lavril, Lee, Liskovich, Lu, Mao, Martinet, Mihaylov, Mishra, Molybog, Nie, Poulton, Reizenstein, Rungta, Saladi, Schelten, Silva, Smith, Subramanian, Tan, Tang, Taylor, Williams, Kuan, Xu, Yan, Zarov, Zhang, Fan, Kambadur, Narang, Rodriguez, Stojnic, Edunov, and Scialom}]{touvron2023llama2}
Hugo Touvron, Louis Martin, Kevin Stone, Peter Albert, Amjad Almahairi, Yasmine Babaei, Nikolay Bashlykov, Soumya Batra, Prajjwal Bhargava, Shruti Bhosale, Dan Bikel, Lukas Blecher, Cristian~Canton Ferrer, Moya Chen, Guillem Cucurull, David Esiobu, Jude Fernandes, Jeremy Fu, Wenyin Fu, Brian Fuller, Cynthia Gao, Vedanuj Goswami, Naman Goyal, Anthony Hartshorn, Saghar Hosseini, Rui Hou, Hakan Inan, Marcin Kardas, Viktor Kerkez, Madian Khabsa, Isabel Kloumann, Artem Korenev, Punit~Singh Koura, Marie-Anne Lachaux, Thibaut Lavril, Jenya Lee, Diana Liskovich, Yinghai Lu, Yuning Mao, Xavier Martinet, Todor Mihaylov, Pushkar Mishra, Igor Molybog, Yixin Nie, Andrew Poulton, Jeremy Reizenstein, Rashi Rungta, Kalyan Saladi, Alan Schelten, Ruan Silva, Eric~Michael Smith, Ranjan Subramanian, Xiaoqing~Ellen Tan, Binh Tang, Ross Taylor, Adina Williams, Jian~Xiang Kuan, Puxin Xu, Zheng Yan, Iliyan Zarov, Yuchen Zhang, Angela Fan, Melanie Kambadur, Sharan Narang, Aurelien Rodriguez, Robert Stojnic, Sergey Edunov, and Thomas
  Scialom. 2023{\natexlab{b}}.
\newblock \href {https://arxiv.org/abs/2307.09288} {Llama 2: Open foundation and fine-tuned chat models}.
\newblock \emph{arXiv preprint arXiv:2307.09288}.

\bibitem[{Wan et~al.(2020)Wan, Wan, and Wang}]{wan-etal-2020-improving}
Zhaohong Wan, Xiaojun Wan, and Wenguang Wang. 2020.
\newblock \href {https://doi.org/10.18653/v1/2020.coling-main.200} {Improving grammatical error correction with data augmentation by editing latent representation}.
\newblock In \emph{Proceedings of the 28th International Conference on Computational Linguistics}, pages 2202--2212, Barcelona, Spain (Online). International Committee on Computational Linguistics.

\bibitem[{Wang et~al.(2022)Wang, Jia, Li, Yu, Xiong, Dong, and Liao}]{10.1145/3510003.3510062}
Deze Wang, Zhouyang Jia, Shanshan Li, Yue Yu, Yun Xiong, Wei Dong, and Xiangke Liao. 2022.
\newblock \href {https://doi.org/10.1145/3510003.3510062} {Bridging pre-trained models and downstream tasks for source code understanding}.
\newblock In \emph{Proceedings of the 44th International Conference on Software Engineering}, ICSE '22, page 287–298, New York, NY, USA. Association for Computing Machinery.

\bibitem[{Xia et~al.(2024)Xia, Malladi, Gururangan, Arora, and Chen}]{xia2024less}
Mengzhou Xia, Sadhika Malladi, Suchin Gururangan, Sanjeev Arora, and Danqi Chen. 2024.
\newblock \href {https://arxiv.org/abs/2402.04333} {Less: Selecting influential data for targeted instruction tuning}.
\newblock \emph{arXiv preprint arXiv:2402.04333}.

\bibitem[{Yannakoudakis et~al.(2018)Yannakoudakis, Andersen, Geranpayeh, Briscoe, and Nicholls}]{yannakoudakis2018developing}
Helen Yannakoudakis, {\O}istein~E Andersen, Ardeshir Geranpayeh, Ted Briscoe, and Diane Nicholls. 2018.
\newblock \href {https://www.tandfonline.com/doi/citedby/10.1080/08957347.2018.1464447?scroll=top&needAccess=true&role=tab} {Developing an automated writing placement system for esl learners}.
\newblock \emph{Applied Measurement in Education}, 31(3):251--267.

\bibitem[{Yannakoudakis et~al.(2011)Yannakoudakis, Briscoe, and Medlock}]{yannakoudakis-etal-2011-new}
Helen Yannakoudakis, Ted Briscoe, and Ben Medlock. 2011.
\newblock \href {https://aclanthology.org/P11-1019} {A new dataset and method for automatically grading {ESOL} texts}.
\newblock In \emph{Proceedings of the 49th Annual Meeting of the Association for Computational Linguistics: Human Language Technologies}, pages 180--189, Portland, Oregon, USA. Association for Computational Linguistics.

\bibitem[{Yuan and Briscoe(2016)}]{yuan-briscoe-2016-grammatical}
Zheng Yuan and Ted Briscoe. 2016.
\newblock \href {https://doi.org/10.18653/v1/N16-1042} {Grammatical error correction using neural machine translation}.
\newblock In \emph{Proceedings of the 2016 Conference of the North {A}merican Chapter of the Association for Computational Linguistics: Human Language Technologies}, pages 380--386, San Diego, California. Association for Computational Linguistics.

\bibitem[{Zhang et~al.(2018)Zhang, Kumar, Khayrallah, Murray, Gwinnup, Martindale, McNamee, Duh, and Carpuat}]{Zhang2018AnEE}
Xuan Zhang, Manish Kumar, Huda Khayrallah, Kenton Murray, Jeremy Gwinnup, Marianna~J. Martindale, Paul McNamee, Kevin Duh, and Marine Carpuat. 2018.
\newblock \href {https://api.semanticscholar.org/CorpusID:53295888} {An empirical exploration of curriculum learning for neural machine translation}.
\newblock \emph{ArXiv}, abs/1811.00739.

\bibitem[{Zhang et~al.(2023)Zhang, Cui, Cai, Huang, Fang, and Bi}]{zhang2023multitask}
Yue Zhang, Leyang Cui, Deng Cai, Xinting Huang, Tao Fang, and Wei Bi. 2023.
\newblock \href {https://arxiv.org/abs/2305.13225} {Multi-task instruction tuning of llama for specific scenarios: A preliminary study on writing assistance}.
\newblock \emph{arXiv preprint arXiv:2305.13225}.

\bibitem[{Zhang et~al.(2022)Zhang, Li, Bao, Li, Zhang, Li, Huang, and Zhang}]{zhang-etal-2022-mucgec}
Yue Zhang, Zhenghua Li, Zuyi Bao, Jiacheng Li, Bo~Zhang, Chen Li, Fei Huang, and Min Zhang. 2022.
\newblock \href {https://doi.org/10.18653/v1/2022.naacl-main.227} {{M}u{CGEC}: a multi-reference multi-source evaluation dataset for {C}hinese grammatical error correction}.
\newblock In \emph{Proceedings of the 2022 Conference of the North American Chapter of the Association for Computational Linguistics: Human Language Technologies}, pages 3118--3130, Seattle, United States. Association for Computational Linguistics.

\bibitem[{Zhao et~al.(2020)Zhao, Wu, Xu, Tao, Zhao, and Yan}]{zhao-etal-2020-knowledge-grounded}
Xueliang Zhao, Wei Wu, Can Xu, Chongyang Tao, Dongyan Zhao, and Rui Yan. 2020.
\newblock \href {https://doi.org/10.18653/v1/2020.emnlp-main.272} {Knowledge-grounded dialogue generation with pre-trained language models}.
\newblock In \emph{Proceedings of the 2020 Conference on Empirical Methods in Natural Language Processing (EMNLP)}, pages 3377--3390, Online. Association for Computational Linguistics.

\bibitem[{Zhou et~al.(2020{\natexlab{a}})Zhou, Ge, Mu, Xu, Wei, and Zhou}]{zhou-etal-2020-improving-grammatical}
Wangchunshu Zhou, Tao Ge, Chang Mu, Ke~Xu, Furu Wei, and Ming Zhou. 2020{\natexlab{a}}.
\newblock \href {https://doi.org/10.18653/v1/2020.findings-emnlp.30} {Improving grammatical error correction with machine translation pairs}.
\newblock In \emph{Findings of the Association for Computational Linguistics: EMNLP 2020}, pages 318--328, Online. Association for Computational Linguistics.

\bibitem[{Zhou et~al.(2020{\natexlab{b}})Zhou, Yang, Wong, Wan, and Chao}]{zhou-etal-2020-uncertainty}
Yikai Zhou, Baosong Yang, Derek~F. Wong, Yu~Wan, and Lidia~S. Chao. 2020{\natexlab{b}}.
\newblock \href {https://doi.org/10.18653/v1/2020.acl-main.620} {Uncertainty-aware curriculum learning for neural machine translation}.
\newblock In \emph{Proceedings of the 58th Annual Meeting of the Association for Computational Linguistics}, pages 6934--6944, Online. Association for Computational Linguistics.

\end{thebibliography}

\clearpage
\appendix

\section{Appendix}
\label{sec:appendix}
\subsection{Hyper-parameter settings for fine-tuning LLaMA2}\label{appendix:parameters}
Table~\ref{Tab:papameters} presents the hyper-parameter settings for supervised fine-tuning LLaMA2 models. 
\begin{table}[ht]
\small
\centering
\resizebox{.40\textwidth}{!}{
\begin{tabular}{l l}
\toprule
\textbf{Hyper-parameters} & \textbf{Settings} \\
\cmidrule(r){1-2}
Learning Rate & 1e-6 \\
Min-lr & 9e-7 \\
Learning Rate Scheduler & Cosine \\
Weight Decay & 0.1 \\
Epoches & 3 \\
Seed & 42 \\
Seq-length & 4096 \\
Batch Size Per GPU & 1 \\
Gradient Accumulation Steps & 4 \\
Pipeline Model Parallel Size & 4  \\
Tensor Model Parallel Size & 2 \\
Cross-doc & yes \\
Attention Dropout & 0.3 \\
Hidden Dropout & 0.3 \\
Output Dropout & 0.3 \\
\bottomrule
\end{tabular}}
\caption{Hyper-parameters for supervised fine-tuning LLaMA2 on English GEC data.}
\label{Tab:papameters}
\end{table}

\subsection{Examples of Curriculum Ratings from LLaMA2-70b}\label{appendix:examples}
\begin{figure*}[ht]
\centering
\includegraphics[width=0.98\textwidth, trim=0 0 0 0]{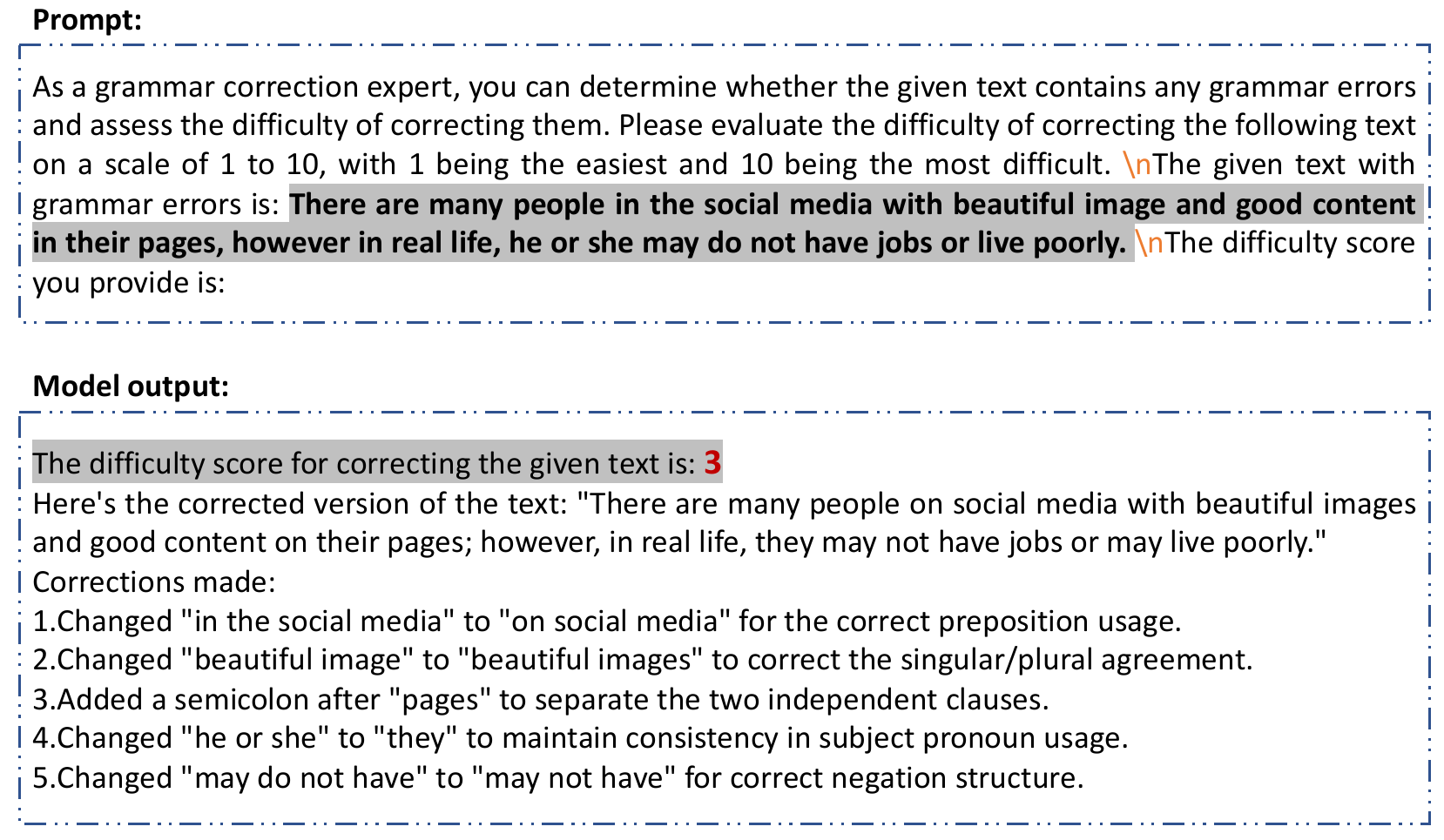}
\caption{The designed scoring prompts for the LLaMA2-70b model and evaluated the difficulty of correcting incorrect sentences provided by the model.}
\label{fig:exampls}
\end{figure*}

Figure \ref{fig:exampls} shows the prompts we designed for LLaMA2-70b and the scores generated by the model. Note that the model does not directly provide scores during generation. After collecting the model's responses, we performed post-editing to derive the final numerical scores.

\subsection{CL Performance at Different Stages}\label{appendix:CL_pre_ds}
\begin{figure}[!t]
\centering
\includegraphics[width=0.48\textwidth, trim=0 0 0 0]{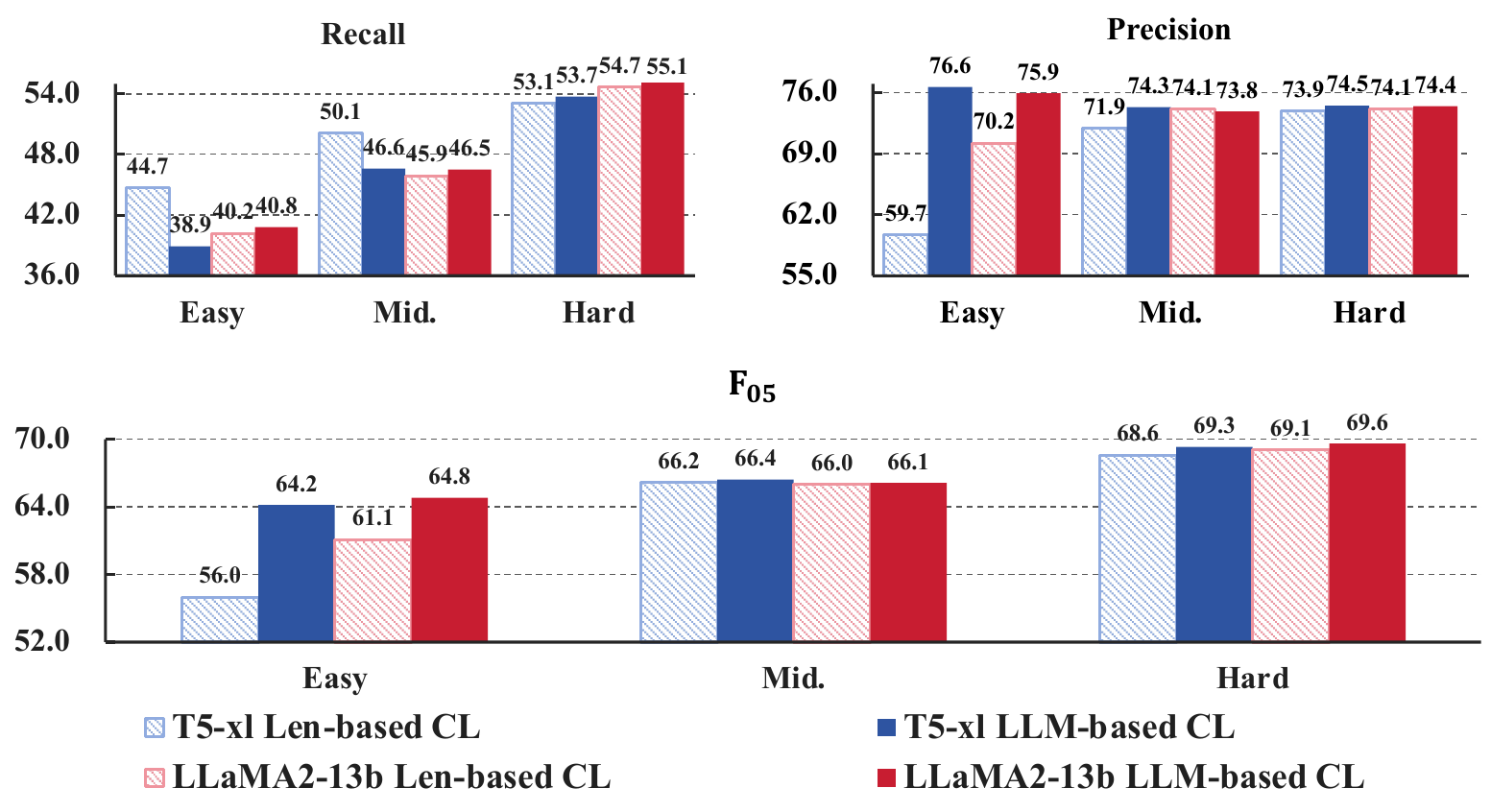}
\caption{Progression of Recall, Precision, and F$_{0.5}$ from Easy to Hard stages on CoNLL14 test set using Len-based CL and LLM-based CL methods on T5-xl and LLaMA2-13b.}
\label{fig:easy2hard}
\end{figure}

To investigate the performance at different stages of the CL method, we examined the progression of \textbf{Recall}/ \textbf{Precision}/ \textbf{F$_{0.5}$} from Easy to Hard stages. We selected the Len-based CL and LLM-based CL methods implemented on T5-xl and LLaMA2-13b for analysis on the CoNLL14 test set. Figure~\ref{fig:easy2hard} presents the detailed results.
During the initial Easy stage, the performance of T5-xl and LLaMA2-13b demonstrated inconsistency due to the varying criteria set by the Len-based and LLM-based CL methods. The Precision and F$_{0.5}$ scores of the LLM-based CL method significantly outperformed those of the Len-based CL in both models. However, the Len-based CL method in T5-xl yielded superior Recall results over the LLM-based CL, whereas the performance was indifferent in LLaMA2-13b. 
As we progressed to the Medium difficulty stage, the two models exhibited contrasting performances with their two methods in terms of precision and recall scores. The Len-based method on T5-xl surpassed the LLM-based method, while the reverse was true for LLaMA2-13b. This discrepancy may not solely be attributed to the differences in model architecture, but could also be due to the LLM-based CL method utilizing a larger LLaMA2 series model (-70b) to select data based on difficulty level. This approach might be more conducive to student models structured around the LLaMA2 framework.
Upon reaching the hard stage, we found that our proposed LLM-based CL method consistently excelled over the Len-based CL method, regardless of whether we were evaluating based on Recall, Precision, or the composite F$_{0.5}$ score. 
This suggests that utilizing a more powerful LLM to assess difficulty levels and sequentially training the model from easy to hard data is a more effective strategy. This is further evidenced by the improvement in F$_{0.5}$ scores from the Easy to Hard stages.

\begin{table*}[ht]
\small
\centering
\resizebox{0.98\textwidth}{!}{
\begin{tabular}{l|rrr|rrr|rrr|rrr} 
\toprule
{\multirow{2}{*}
{ \textsc{\textbf{Method}}}}  & 
\multicolumn{3}{c|}{\textsc{\textbf{R (61.4\%)}}} &
\multicolumn{3}{c|}{\textsc{\textbf{M (26.9\%)}}} &
\multicolumn{3}{c|}{\textsc{\textbf{U (10.4\%)}}} &
\multicolumn{3}{c}{\textsc{\textbf{WO (1.3\%)}}} \\
\cmidrule(r){2-13}
& Pre. & Rec. & F$_{0.5}$ & Pre. & Rec. & F$_{0.5}$ & Pre. & Rec. & F$_{0.5}$ & Pre. & Rec. & F$_{0.5}$ \\
\cmidrule{1-13}
T5-large GEC & 61.2 & 42.9 & 56.4 & 59.0 & 44.5 & 55.4 &60.3 & 40.6 & 55.0 & 48.0 & 37.9 & 45.6 \\
T5-large Len-based CL & 61.5 & 43.4 & \bf 56.8 & 59.4 & 44.0 & 55.5 & 59.6 & 42.9 & 55.3 & 46.8 & 37.9 & 44.7 \\
T5-large LLM-based CL & 61.1 & 43.6 &  56.5 & 60.1 & 46.0 & \bf 56.7 & 60.1 & 42.8 & \bf 55.6 & 48.7 & 39.0 & \bf 46.4 \\
\cmidrule{1-13}
T5-xl GEC & 62.4 & 45.1 & 58.0 & 60.9 &45.6 &57.1  & 61.7 &43.1 &56.8 & 46.8 & 37.9 & 44.7 \\
T5-xl Len-based CL & 62.2 & 45.3 & 57.9 & 61.7 & 46.4 & \bf 57.9 & 62.0 & 42.9 & 56.9 & 47.2 & 35.8 & 44.4 \\
T5-xl LLM-based CL & 63.5 & 44.8 & \bf 58.6 & 62.0 & 44.4 & 57.4 & 61.8 & 43.3 & \bf 57.0 & 48.0 & 37.9 & \bf 45.6 \\
\cmidrule{1-13}
LLaMA2-7b GEC & 61.6 & 43.2 & 56.8 & 57.7 & 40.6 & 53.2 & 59.5 & 39.8 & 54.1 & 46.9 & 31.6 &  42.8 \\
LLaMA2-7b Len-based CL & 61.2 & 44.3 & 56.9 & 57.4 & 43.7 & 54.0 & 59.5 & 41.5 & 54.7 & 50.0 & 34.7 & 46.0 \\
LLaMA2-7b LLM-based CL &60.9 & 45.1 & \bf 56.9 & 57.7 & 44.6 & \bf 54.5 & 58.9 & 43.3 & \bf 54.9 & 51.5 &36.8 & \bf47.7 \\
\cmidrule{1-13}
LLaMA2-13b GEC & 63.8& 41.6 & 57.6 & 64.5 & 36.6 & 56.0 & 64.5 & 36.6 & 56.0 & 50.0 & 28.4& 43.4 \\
LLaMA2-13b Len-based CL & 61.9 & 46.1 & 58.0 & 59.0 & 48.7 & \bf 56.6 & 61.3 & 43.3 & 56.6 & 48.6 & 35.8 & 45.4 \\
LLaMA2-13b  LLM-based CL & 61.8 & 46.4 & \bf 58.0 & 58.9 & 47.7 &  56.2 & 61.0 & 43.9 & \bf 56.6 & 51.4 & 37.9 & \bf 48.0 \\
\bottomrule
\end{tabular}}
\linespread{1}
\caption{
The performance of the T5-large GEC/ Len-based CL/ LLM-based CL, and LLaMA2-13b GEC/ Len-based CL/ LLM-based CL models on Operation-Level error types is showcased using the BEA19 development set. 
The values in parentheses denote the proportion of each operation-level error type. The highest F$_{0.5}$ scores are highlighted in \textbf{bold} in each category. 
} 
\label{Tab:analy_umro}
\end{table*}
\subsection{Operation-Level Error Types}\label{appendix:error_type_ol}
We evaluate the performance of various Operation-Level error types using the BEA19 development set.
With the help of the ERRANT toolkit, we categorize these errors into four distinct categories: Replacement, Missing, Unnecessary, and Word Order (WO). 
Followed by \citet{fang-etal-2023-improving}, we manual distinguish  Word Order from the Replacement category.
Table~\ref{Tab:analy_umro} reveals the performance of different models in handling four types of errors under the baseline model, the len-based CL method, and our proposed LLM-based CL method. 
It is observed that both CL methods significantly outperform the baseline model in dealing with these four types of errors. 
In particular, the LLM-based CL method excels the Len-based CL method in dealing with ``Unnecessary'' and ``Word Order'' errors. 
Moreover, it surpasses the length-based CL method in handling ``Replacement'' and ``Missing'' words errors most of the time. 
These results highlight the advantages of our proposed CL method in handling different types of errors, especially its strong capability in dealing with complex error types.

\end{document}